\documentclass{article}

\usepackage[preprint]{neurips_2026}

\usepackage[utf8]{inputenc} 
\usepackage[T1]{fontenc}    
\usepackage{hyperref}       
\usepackage{url}            
\usepackage{booktabs}       
\usepackage{amsfonts}       
\usepackage{nicefrac}       
\usepackage{microtype}      
\usepackage{xcolor}         
\usepackage{amsmath} 
\usepackage{graphicx}
\usepackage{subcaption}

 \usepackage{xcolor}
\newif\ifshowcolors
\showcolorstrue

\ifshowcolors

\else

\fi

\title{PhyWorld: Physics-Faithful World Model for Video Generation}

\author{%
Pu Zhao\thanks{These authors contributed equally.}$\ \ ^1$, Juyi Lin$^{*1}$, Timothy Rupprecht$^1$, Arash Akbari$^1$, \\ \textbf{Chence Yang$^2$, Rahul Chowdhury$^1$, Elaheh Motamedi$^1$, Arman Akbari$^1$,} \\  \textbf{Yumei He$^3$, Chen Wang$^4$, Geng Yuan$^2$, Weiwei Chen$^4$, Yanzhi Wang$^1$} \\
$^1$Northeastern University,  $^2$University of Georgia,   $^3$Tulane University, $^4$EmbodyX \\
  \texttt{\{p.zhao, lin.juy, yanzhiwang\}@northeastern.edu} \\
  \\
PhyWorld: \textit{https://huggingface.co/NU-World-Model-Embodied-AI/phyworld}
}

\begin{document}

\maketitle

\begin{abstract}
World simulators can provide safe and scalable environments for training Physical AI systems before real-world deployment. Large video generation models are emerging as a promising basis for such simulators because they can generate diverse and realistic visual futures. However, using them as world simulators requires physically faithful video continuations, namely, generated videos that preserve the physical state implied by the conditioning input, and evolve in ways consistent with basic physical principles. We propose \textbf{PhyWorld}, a video generation world model designed to produce temporally coherent and physically faithful scene continuations through two-stage post-training. In the first stage, we improve video-to-video continuation with flow matching fine-tuning, encouraging stable visual attributes and coherent motion dynamics across frames. In the second stage, we align generated dynamics with physical principles using Direct Preference Optimization (DPO) over physics preference pairs, guiding the model toward outputs with higher physical plausibility. To evaluate PhyWorld, we use both standard video-quality benchmarks and a dedicated physical-faithfulness benchmark with per-law scoring. Experiments show that PhyWorld improves video consistency, achieving an average score of 0.769 on VBench compared with 0.756 or below for state-of-the-art baselines. PhyWorld also improves physical plausibility, reaching an average score of 3.09 on our physical-faithfulness benchmark compared with 2.99 for the strongest baseline. These results suggest that post-training large video generation models with continuation and physics-preference signals can make them more effective world simulators for Physical AI. 
\end{abstract}

\section{Introduction}

Physical AI systems are embodied agents that perceive the world through sensors and act on it through actuators. Training these agents directly in the real world is slow, expensive, and risky, especially when early-stage policies may produce unsafe or damaging actions. This makes simulation central to Physical AI: before acting in the real world, an agent should be able to practice in environments that are visually rich, diverse, and physically plausible. Large video generation models offer a promising basis for such simulators. Trained on Internet-scale video corpora, models such as Sora, CogVideoX, Veo, Cosmos, and Wan \citep{liu2024sora, yang2024cogvideox, wiedemer2025video, agarwal2025cosmos, wan2025wan} learn rich visual priors about how objects move, collide, fall, deform, and interact with light. Unlike classical physics engines, which often require hand-authored assets, materials, and rules for each domain, video generation models can synthesize diverse scene continuations from text, images, or preceding clips. This makes video-to-video continuation especially relevant for world simulation: given an observed scene, the model must generate what happens next while preserving the physical state implied by the input. We therefore study video generation models as world simulators for Physical AI. Specifically, we ask whether they can generate physically faithful scene continuations: videos that remain temporally consistent across frames and whose dynamics are plausible under basic physical principles.

Existing video generation world models exhibit two fundamental challenges that undermine their capacity to generate physically plausible video content. First, the physical consistency is hard to maintain across generated frames. As a result, 
the generation  quality of   state-of-the-art models in this domain remains limited. Artifacts such as background color drift and inconsistent object motion speeds are frequently observed, undermining temporal coherence and visual realism. Text-to-video and image-to-video approaches are similarly inadequate, as neither modality provides sufficient grounding for inferring fine-grained physical attributes from static or linguistic inputs alone.
Second, physical law enforcement is largely absent from existing model architectures. Prevailing world models are trained on empirical video data without incorporating explicit supervisory signals or loss functions designed to instill knowledge of physical principles. Consequently, these models lack a structured mechanism for internalizing laws governing dynamics, causality, or physical interactions. This architectural gap manifests in generated outputs that may violate fundamental physical constraints, rendering the models both unaware of and non-compliant with the underlying laws that govern realistic physical environments.

To address the above  challenges,  we propose our video-generation world model, PhyWorld, to generate more consistent and physically faithful videos. It includes two stage training: (1)  physical consistency  enhancement, and (2)  physics enforcement through reinforcement learning.   The first stage targets to enable the video-to-video generation capability and  improve the video continuation performance through flow matching fine-tuning, thus continuing  the  scene with consistent physical behavior (such as color and speed of objects).  Besides achieving superior performance on generation quality, the second stage targets to enforce physics   through reinforcement learning, with explicit supervisory signals or loss functions designed to instill knowledge of physical principles. 
This stage is based on evaluating the physics faithfulness of the PhyWorld model using 
our 250-prompt text/image-to-video benchmark organized under a taxonomy of physical laws. Each generated video is scored on a 1-5 Likert scale across general quality dimensions and physics-specific dimensions using our developed 9B open-weight video-language judge model. Our evaluation demonstrates the outstanding performance on physical consistency and plausibility. Our paper makes three key contributions. 

\begin{itemize}
    \item \textbf{Physical Consistency  Enhancement.} 
     We introduce a video-to-video training pipeline to effectively produce temporally coherent and visually stable continuations. It substantially improves the temporal consistency during video continuation. 
    \item \textbf{Physics Enforcement with Reinforcement Learning and Benchmarking.}
To further align generated video content with real-world physical principles, we employ Direct Preference Optimization (DPO) to finetune the model obtained from the first training stage. This approach incentivizes the model to preferentially generate outputs that adhere to physically plausible dynamics, thereby improving compliance with fundamental physical laws in practice. This DPO-based reinforcement learning process leverages our text/image-to-video physics benchmark to evaluate the physical plausibility of generated videos.
    \item \textbf{Superior Performance on Physical Evaluation.} 
   Our evaluation demonstrates that PhyWorld  can achieve superior generation performance with outstanding video generation quality  (an average score of 0.769 v.s. 0.756 or below from SOTA baselines on VBench \cite{huang2023vbench})  and physics alignment  (an average score of 3.09 v.s. 2.99 from SOTA baselines on our dedicated physical faithfulness benchmark). 
\end{itemize}

\section{Related Work}

\subsection{World Model}
A synthesis of recent surveys on state-of-the-art world models \cite{ding2025understanding, li2025comprehensive, yue2025simulating, long2025survey,zhao2025open,zhao20247b, lin2025exploring, liu2025generative, xu2026specialist, dong2026learning} reveals a consistent architectural paradigm: to support planning and decision-making (particularly in embodied settings \cite{long2025survey, li2025comprehensive}), world models are broadly designed to function as simulators that (1) represent the current state of the world and (2) predict future world dynamics \cite{ding2025understanding}.
Recent literature has further extended this paradigm across several specialized dimensions, including advances in video-based world models \cite{yue2025simulating}, embodied intelligence \cite{li2025comprehensive}, joint temporal-spatial modeling \cite{liu2025generative}, and physical realism \cite{lin2025exploring}. Collectively, these works converge on a set of persistent open challenges, notably long-horizon temporal consistency, computational scalability, and faithful alignment with the governing principles of real-world physics.

The landscape of world models is organized  along several distinct axes, including domain-specific applications \cite{long2025survey, yue2025simulating, li2025comprehensive, shen2025iccad, add-zhao-etal-2024-pruning,shen2024search,shen2025quart,lin2025exploring, dong2026learning}, architectural and input modality distinctions \cite{liu2025generative}, and abstract taxonomic frameworks \cite{ding2025understanding, xu2026specialist}. 
Beyond the established contributions from SOTA world models such as V-JEPA \cite{hansen2024hierarchical, garrido2026learning, yuan2026inference},  Wan \cite{wan2025wan}, Cosmos  \cite{kim2026cosmos} and PointWorld \cite{yang2025cambrian, wang2025vagen, huang2026pointworld}, a growing body of work explores architectural innovations \cite{shen2025sparse,zhan2024exploring,zhan-etal-2024-rethinking-token,10.1145/3418297,mi2026effective} that push the boundaries of world model design, including Vision-Action models \cite{li2026causal}, Vision-Language-Action frameworks \cite{team2026advancing,lin2025vote}, autoregressive  (AR) architectures \cite{gao2025adaworld,shen2026fastcar,shen2025numerical,shen2025lazydit}, and diffusion-based approaches \cite{zhang2025epona,zhao2026hieramp,zhao2025taming,NEURIPS2024_18b0b4c7,li2025dawm}. Nevertheless, a unified world model framework that reconciles these diverse paradigms remains an open and elusive goal.

Several recent efforts have explored AR video generation as a means of enabling temporally extended world simulation. Self-Forcing \cite{huang2025selfforcing,cui2025self} addresses the training-inference discrepancy in AR video generation by training the model to condition on its own previously generated frames rather than ground-truth frames, thereby reducing compounding errors during long-horizon rollouts. Similarly, LongCAT \cite{meituanlongcatteam2025longcatvideotechnicalreport,yang2025longlive,team2026longcat} proposes a framework for generating long, temporally consistent video sequences through a context-aware tokenization strategy.
However, these methods are relatively compact in terms of model scale and consequently exhibit notable limitations in generation quality, including reduced visual fidelity, weaker photorealism, and less reliable physical consistency compared to larger-scale video generation models. This trade-off between model size and generation quality represents a fundamental constraint for their use as world simulators in Physical AI training pipelines, where both temporal coherence and physical  adherence are critical requirements.

\subsection{Physics Evaluation}

A recent line of work has sought to systematically evaluate the degree to which generated videos adhere to physical laws. However, each existing benchmark embeds methodological choices that constrain its discriminative power in important ways.
\textsc{Physics-IQ} \citep{motamed2026generative} introduces pixel-level reference metrics (including spatial IoU, spatiotemporal IoU, and mean squared error) across 66 image-to-video scenarios. However, its design pre-supposes a unique ground-truth trajectory, thereby penalizing both legitimate camera motion and physically valid stochastic outcomes such as collision rebound angles and fluid splash dynamics. 
\textsc{PhyGenBench} \citep{meng2024towards} extends coverage to 160 prompts spanning 27 physical laws, yet relies on a three-stage cascaded binary scoring mechanism in which classification errors propagate across successive stages, potentially compounding evaluation noise.
\textsc{VideoPhy-2} \citep{bansal2025videophy} scales to approximately 590–688 prompts with 12 human annotators, but reports only two coarse evaluation axes (semantic adherence and holistic physical correctness) without providing per-law decomposition, limiting the granularity of physical assessment.
\textsc{WorldModelBench} \citep{li2025worldmodelbench} encompasses 350 prompts but evaluates only 5 physical laws using binary per-law judgments.

A synthesis of the limitations described above reveals three cross-cutting methodological failures that recur across existing benchmarks with more details in Appendix~\ref{app:sec:failure}.
(i) A pervasive reliance on implicit prompts,  which describe a physical scene without explicitly specifying the expected outcome, means that evaluators can only assess whether the generated video appears visually plausible, rather than whether the underlying physics is demonstrably correct. In the absence of a well-defined physical expectation, evaluation necessarily collapses into a subjective judgment of surface-level realism.
(ii) The predominant use of holistic or binary scoring rubrics obscures concrete physical law violations within a single aggregate score, precluding the kind of per-law diagnostic analysis necessary to identify and interpret model-specific failure modes. Such coarse-grained scoring conflates qualitatively distinct physical errors, rendering benchmarks unable to attribute performance differences to specific physical phenomena.
(iii) A widespread dependence on closed-source evaluators (including GPT-4o, GPT-5, and Gemini-3.1-Pro) introduces compounding reliability concerns. The API behavior of these models is subject to silent drift over time, and empirical inspection reveals systematic failure  on specific physics domains. Notably, Gemini-3.1-Pro has been observed to  classify visibly dynamic fluid motion as completely static, and produces markedly unstable scores on shadow and reflection phenomena. 

\section{Methodology}

\begin{figure}[t]
    \centering
    \includegraphics[width=0.94\textwidth]{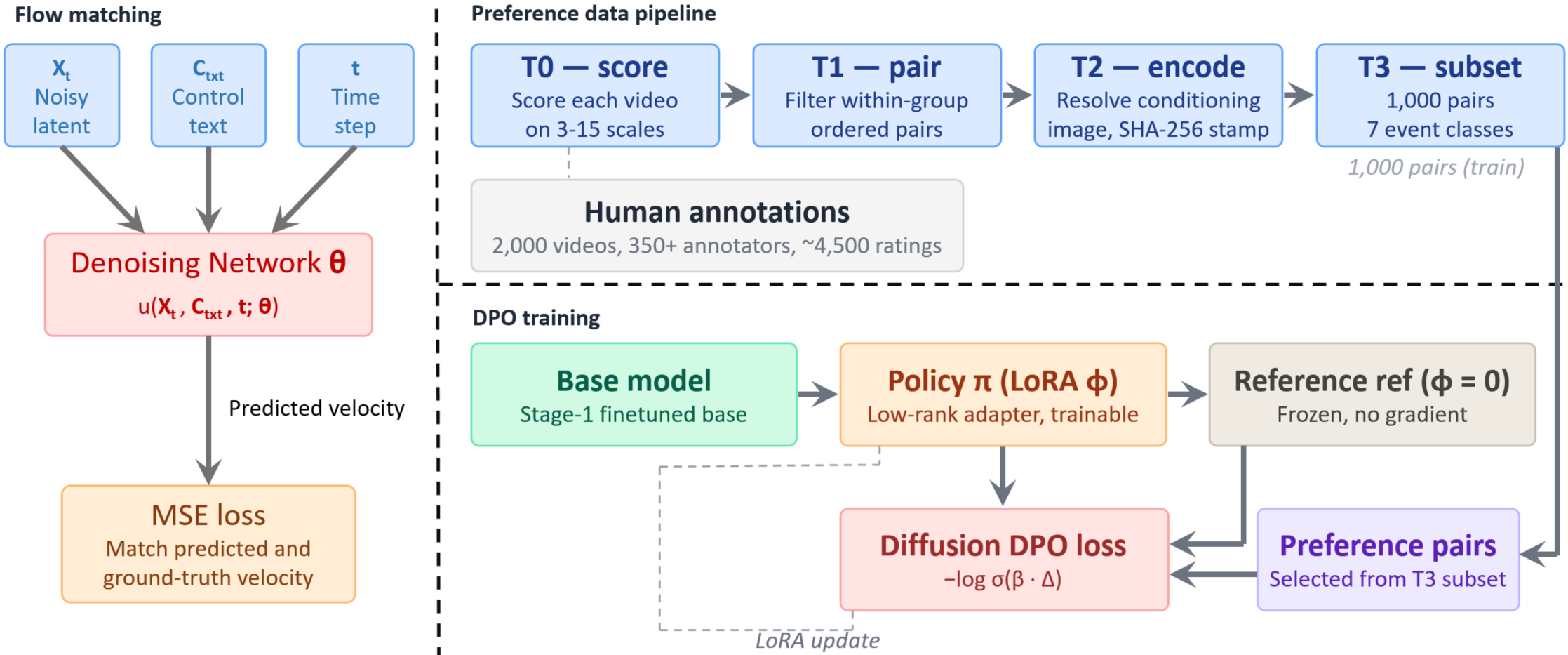}
    \caption{Framework overview with flow matching and DPO training. We first finetune our model to enhance the physical consistence through flow matching. Then we  adopt  DPO  to train  the model by enforcing physics, with explicit supervisory signals   to instill knowledge of physical principles.}
    \label{fig:system}
\end{figure}

We first finetune our model to enhance the physical consistence through flow matching training. Then we  adopt Direct Preference Optimization (DPO) \cite{rafailov2023direct}  to train  the model by enforcing physics, with explicit supervisory signals or loss functions designed to instill knowledge of physical principles.  We demonstrate our framework in Figure~\ref{fig:system}.

\subsection{Physical Consistence Enhancement}

\subsubsection{Video-to-Video Generation}

We introduce a video prompt as an additional conditioning signal to guide video synthesis. Specifically, the conditioning video $I \in \mathbb{R}^{C \times R \times H \times W}$, consisting of $R$ frames, is concatenated with zero-padded frames along the temporal axis to form the guidance tensor $I_c \in \mathbb{R}^{C \times T \times H \times W}$, which is subsequently compressed by the Wan-VAE encoder \cite{wan2025wan}  into a condition latent representation $z_c \in \mathbb{R}^{c \times t \times h \times w}$, where $c = 16$ denotes the latent channel dimensionality, $t = 1 + (T-1)/4$, $h = H/8$, and $w = W/8$. To explicitly distinguish between preserved and generated frames, we further introduce a binary mask $M \in \{0, 1\}^{1 \times T \times h \times w}$, where a value of 1 designates frames to be retained from the conditioning input and 0 designates frames to be synthesized. The spatial dimensions of $M$ are consistent with those of the condition latent $z_c$. However, $M$ adopts the same   the temporal length of the target video and is subsequently rearranged to the shape $s \times t \times h \times w$, where $s$ denotes the temporal stride of the Wan-VAE. The noise latent $z_t$, the condition latent $z_c$, and the rearranged mask $m$ are then concatenated along the channel axis and fed into the Wan DiT model for denoising.

To further enrich the conditioning signal with global semantic context, we employ the CLIP encoder~\cite{radford2021learning} to extract feature representations from the final frame of the conditioning video. The extracted features are subsequently projected into the model's feature space via a three-layer multi-layer perceptron (MLP), whose output serves as a global context embedding. This embedding is injected into the DiT model through a decoupled cross-attention mechanism, enabling the model to leverage high-level semantic cues throughout the generation process. Together, the proposed masking mechanism and global context injection explicitly delineate the boundary between conditioning input and generation targets, providing structured and semantically grounded guidance for faithful video continuation and enabling effective video-to-video generation.

\subsubsection{Data Processing}

We adopt OpenVid-1M \cite{nan2024openvid} to finetune our world model. OpenVid-1M is a high-quality  video dataset with text descriptions designed for research institutions to enhance video quality, featuring high aesthetics, clarity, and resolution.  

To ensure dataset quality, we apply a multi-stage filtering pipeline designed to retain video clips exhibiting suitable temporal consistency and motion characteristics. For temporal consistency, we employ CLIP \cite{radford2021learning} to extract visual features from consecutive frames and compute pairwise cosine similarity between adjacent frames as a proxy for inter-frame coherence. Clips yielding excessively high similarity scores (indicative of near-static content), as well as those yielding excessively low scores (indicative of frequent flickering or abrupt scene transitions), are discarded, yielding a filtered subset of temporally stable clips.

As temporal consistency alone is insufficient to exclude clips containing high-speed objects that nonetheless maintain frame-to-frame coherence, we additionally incorporate a motion-based filtering criterion. Specifically, we apply UniMatch \cite{unimatch} to estimate optical flow and derive a per-clip motion difference score. Clips exhibiting high optical flow magnitudes, corresponding to rapid or erratic motion, are deemed unsuitable for training and subsequently removed. Clips falling at both extremes of the motion score distribution are filtered out, producing a final subset characterized by smooth, controlled motion that is conducive to stable and consistent video generation.

\subsubsection{Training with Flow Matching}
We adopt the flow matching framework~\cite{lipman2022flow, esser2024scaling} to model a unified generative process applicable to both image and video domains. The training procedure follows a progressive curriculum: we begin with  training on low-resolution videos, followed by  additional optimization over  videos at progressively increasing spatial resolutions and temporal durations. This staged training strategy allows the model to develop robust low-level visual priors before scaling to the full spatiotemporal complexity of high-resolution video generation.

Flow matching provides a theoretically principled foundation for learning continuous-time generative processes within the diffusion framework. Rather than relying on iterative score-based velocity prediction, flow matching parameterizes the generative process as an ordinary differential equation (ODE), enabling stable and efficient training while maintaining equivalence to maximum likelihood estimation objectives. Formally, given an image or video latent $x_1$, a Gaussian noise sample $x_0 \sim \mathcal{N}(\mathbf{0}, \mathbf{I})$, and a timestep $t \in [0, 1]$ drawn from a logit-normal distribution, an intermediate latent $x_t$ is constructed as a linear interpolation along the probability flow trajectory and serves as the training input to the denoising network. 
Following Rectified Flows (RFs) \cite{esser2024scaling}, $x_t$ is defined as a linear interpolation between $x_0$ and $x_1$, i.e.,
\begin{align}
    x_t = t x_1 + (1-t) x_0
\end{align}
The ground-truth velocity $v_t$ is
\begin{align}
   v_t = \frac{d x_t}{d t} =   x_1 - x_0
\end{align}
The model is trained to predict the velocity. Thus, the loss function can be formulated as the mean squared error (MSE) between the model output and $v_t$,
\begin{align}
    \mathcal{L} = \mathbb{E}_{x_0,x_1,c_{txt},t;\theta} \| u(x_t, c_{txt}, t;\theta ) -v_t \|^2
\end{align}
where $c_{txt}$ is the umT5 text embedding sequence of 512 tokens long, $\theta$ is the model weights, and $u(x_t, c_{txt}, t;\theta )$ denotes the output velocity predicted by the model.

\subsection{Physical Law Enforcement with Reinforcement Learning}
\label{sec:rl}

\paragraph{Overview.}
We treat physical law violations (more details in Appendix~\ref{app:sec:failure}) in a pretrained image-to-video (I2V) diffusion model as a preference learning problem and correct them with offline reinforcement learning (RL), specifically direct preference optimization (DPO), over preference pairs. The base model is our derived model from the first stage (starting from the Wan2.2-I2V-A14B model but adapting the model to both image-to-video and video-to-video generation). The policy $\pi$ is the base denoiser augmented with a low-rank LoRA adapter, whose trainable weight increment denoted by $\phi$ (i.e., $\pi$ adds $\phi$ on top of the frozen base weights). It is optimized against a fixed reference $\mathrm{ref}$ that copies the same frozen base weights with $\phi=0$, so the physics-correcting signal does not perturb the base model's conditioning or generation prior.

\paragraph{Data provenance.}
The raw preference signal is taken directly from the  human annotation pool of our text/image-to-video (TI2V) physics benchmark used for evaluation (Sec.~\ref{sec:benchmark}). Concretely, every video that enters the pipeline below is one of the benchmark's $2{,}000$ pre-rated videos ($250$ prompts $\times$ $8$ generators), each scored by multiple human annotators on the $1$--$5$ Likert scale across the three general dimensions (including semantic alignment (SA), physical-temporal validity (PTV), and  persistence), as well as the applicable physics laws, drawn from a quality-controlled multi-annotator study of approximately $350$ raters yielding $\sim$$4{,}500$ cleaned annotations. We reuse these released ratings as-is, without any re-collection, re-judging, or model-derived scoring. The pipeline below only filters, aggregates, and pairs them.

\begin{table}[t]
    \centering
    \begin{minipage}{0.45\textwidth}
        \centering
\caption{T0$\to$T1$\to$T2$\to$T3 preference-pair funnel.
T1 retained excludes 203 ties, 730
low-margin pairs, and 268 rater-filtered pairs;
heldout is evaluation only. T2 is the train$\cup$val candidate pool;
T3 is the RL sampling pool. \textbf{Trainset}
is the 1{,}000-pair round-4 class balanced subset sampled from
T3, with quotas in Table~\ref{tab:round4-quota}.}
\label{tab:pipeline-funnel}
\begin{tabular}{lrrr}
\hline
Stage / pool & \#pairs & \#groups & \#prompts \\
\hline
T1 retained & 3{,}324 & 250 & 253 \\
\quad heldout & 579 & 245 & 42 \\
T2 cand. pool & 2{,}745 & \textemdash & 211 \\
T3 RL pool & 2{,}202 & 208 & \textemdash \\
\textbf{Trainset} & 1{,}000 & \textemdash & \textemdash \\
\hline
\end{tabular}
    \end{minipage}
    \hfill 
    \begin{minipage}{0.48\textwidth}
        \centering
\caption{Round-4 trainset class quota. The 1{,}000-pair trainset
 is sampled from   T3   pool (2{,}202 pairs)
 with   per-class budget below, fixed in the manifest before any
 score from this run is read.}
\label{tab:round4-quota}
\begin{tabular}{lr}
\hline
Physical-event class & \#pairs \\
\hline
A (collision/rebound) & 513 \\
B (destruction/deformation) & 93 \\
C (fluids/liquids) & 168 \\
D (shadow/reflection) & 68 \\
E (chain/multi-stage) & 13 \\
F (rolling/sliding) & 75 \\
G (throwing/ballistic) & 55 \\
unclassified & 15 \\
\hline
\textbf{total} & \textbf{1{,}000} \\
\hline
\end{tabular}
    \end{minipage}
\end{table}

\paragraph{Preference data pipeline (T0$\to$T1$\to$T2$\to$T3).}
Pairs are produced by a four-stage, content-addressed offline pipeline (Table~\ref{tab:pipeline-funnel}). \textbf{T0} fixes a per-video aggregate quality score $s(v) = \tfrac{1}{|\mathcal{R}(v)|}\sum_{r\in\mathcal{R}(v)}(\mathrm{SA}_r(v) + \mathrm{PTV}_r(v) + \mathrm{persistence}_r(v))$ on a $3$--$15$ summed-axis scale, where $r$ indexes individual human raters, $\mathrm{SA}_r(v)$, $\mathrm{PTV}_r(v)$, $\mathrm{persistence}_r(v) \in \{1,\dots,5\}$ are the three axis ratings given by rater $r$ to video $v$, and $\mathcal{R}(v)$ is the cross-group union of complete-triple raters for $v$ (a within-group mean is degenerate because at most one rater is assigned per (group, video) cell). \textbf{T1} enumerates within-group ordered pairs $(v_w, v_l)$, where $v_w$ (winner) is preferred to $v_l$ (loser). A pair is admitted only if the score margin satisfies $s(v_w)-s(v_l) \ge 1.0$ and each video has at least $r_{\min}=2$ raters.  Cross-prompt and cross-group pairings abort the run. The export retains $3{,}324$ pairs over $250$ unique (prompt, physical-law) groups, with a prompt-disjoint $0.7/0.15/0.15$ split into $171/40/42$ prompts. The $42$-prompt heldout slice ($579$ pairs) is the sole evaluation set and is verified absent from every encode-cache and latent sidecar produced downstream. We refer to the remaining train$\cup$val portion ($2{,}745$ pairs over $211$ prompts) as \emph{Tier B-candidate} -- the RL candidate pool that subsequent stages further filter. \textbf{T2} resolves and pre-encodes the I2V conditioning image, drops pairs whose first frame is missing on disk (the surviving $2{,}202$ pairs over $208$ groups form \emph{Tier B-cond-present}), and stamps a SHA-256 over the bytes actually fed to the model, so that host-local fallback resolution cannot silently swap images. \textbf{T3} packages explicitly named subsets from \emph{Tier B-cond-present}. The reported run uses a $1{,}000$-pair, class-balanced round-4 subset over seven induced physical event classes (collision/rebound, destruction/deformation, fluids, shadow/reflection, chain, rolling/sliding, throwing/ballistic, taxonomy verified $42/42$ on heldout). Pair budget, per-class quotas (Table~\ref{tab:round4-quota}), and seed are fixed in the manifest before any score is read.

\paragraph{DPO objective.}
For a winner/loser video pair $(w, l)$ (shorthand for $(v_w, v_l)$ from T1) sharing the same text prompt $c$ and conditioning image $x_0$ (the I2V first frame), let $\mathrm{MSE}_{\theta,v}$ denote the noise prediction mean-squared error of a denoiser with parameters $\theta$ on video $v$, evaluated at diffusion timestep $t \sim \mathcal{U}[t_{\min},t_{\max}]$ with a Gaussian noise sample $\epsilon \sim \mathcal{N}(\mathbf{0},\mathbf{I})$ that is shared across $w$ and $l$ (paired-noise variance reduction). The diffusion DPO loss is
\[
\mathcal{L}_{\mathrm{DPO}} = -\log\sigma(\beta\,\Delta), \quad
\Delta = \bigl(\mathrm{MSE}_{\pi,l}-\mathrm{MSE}_{\pi,w}\bigr) - \bigl(\mathrm{MSE}_{\mathrm{ref},l}-\mathrm{MSE}_{\mathrm{ref},w}\bigr),
\]
where $\sigma(\cdot)$ is the logistic sigmoid, $\pi$ is the LoRA-augmented policy and $\mathrm{ref}$ the frozen base. $\beta>0$ is the DPO inverse temperature that controls the strength of the preference signal, and $\Delta$ is the implicit reward margin, i.e., how much more $\pi$ (relative to $\mathrm{ref}$) prefers $w$ over $l$ in noise prediction error. The reference branch is computed without gradient tracking by zeroing the LoRA increment $\phi$ in place (rather than instantiating a second model), halving activation memory at the cost of one extra forward per pair. We restrict training to the high-noise window $t\in[901,999]$ to suppress reward-hacking via timestep selection, matching where the automated judge places most of its discriminative signal.

\paragraph{Policy and training.}
The LoRA increment $\phi$ is parameterized as a low-rank adapter with rank $r{=}16$ and scaling factor $\alpha{=}16$, applied to the attention and feed-forward projection matrices of the Wan2.2 denoiser. All other parameters are frozen. We treat $\beta$ (the DPO inverse temperature defined above) as the only hyper-parameter selected on held-out automated judge validation, sweeping $\beta \in \{30,100,300\}$ and locking $\beta=100$ before unblinding. Under the pre-registered trajectory criterion (axis-averaged Spearman monotonicity plus step-$250$ value), $\beta=100$ at step $250$ wins both metrics (Spearman $+0.520$ vs.\ $+0.020$ for $\beta=30$; final implicit reward margin $\Delta=+0.200$ vs.\ $+0.075$) and is the reported RL checkpoint.

\vspace{-2pt}
\subsection{Benchmark for Physical Faithfulness}
\label{sec:benchmark}

Standard video-quality benchmarks such as VBench~\cite{huang2023vbench} measure visual fidelity and motion smoothness, but are not designed to detect physical law violations. Because the central claim of PhyWorld concerns physical faithfulness, our evaluation needs a benchmark whose scoring is grounded in per-law physical correctness rather than holistic plausibility, and whose protocol distinguishes among categories of physical failure rather than collapsing them into a single aggregate value. 

\paragraph{Benchmark setup.} We evaluate on our text/image-to-video (TI2V) physics benchmark of $250$ prompts, each paired with a conditioning first frame and organized under a structured taxonomy of physical laws. Each generated video is scored on a $1$--$5$ Likert scale along general video-quality dimensions (semantic alignment, physical-temporal validity, persistence) and along the physics-law dimensions (collision/rebound, destruction/deformation, fluids, shadow/reflection, chain, rolling/sliding, and throwing/ballistic) applicable to its prompt. To perform scoring, we finetune the VLM model Qwen3.5-9B \cite{qwen3.5}, released as the video-language judge model. This judge model is 
fine-tuned on a quality-controlled multi-annotator human study (approximately $350$ raters, $\sim$$4{,}500$ cleaned annotations across $13$ physics laws plus the three general dimensions). We use the released judge as-is to give scores for generated videos, so that the resulting scores are reproducible and not subject to closed-source API drift. More details about the benchmark and the judge model are demonstrated in Appendix~\ref{app:sec:more_detail}.
Our benchmark and human annotations are  used not only for  physical faithfulness evaluation, but also to provide preference signals in DPO, as detailed in Sec.~\ref{sec:rl}.

\paragraph{Judge inference.} The released judge model is an open-weight video-language model fine-tuned end-to-end on the human Likert annotations described above. At inference time it takes a frame-sampled clip together with an augmented prompt that explicitly states the expected physical outcome, and emits a single JSON key--value pair containing one dimension name and one integer in $\{1,\dots,5\}$. Crucially, each evaluation dimension---SA, PTV, persistence, and every applicable physics law (including collision/rebound, destruction/deformation, fluids, shadow/reflection, chain, rolling/sliding, and throwing/ballistic)---is queried in a \emph{separate} inference call rather than in a multi-key joint call, so reasoning over one dimension cannot leak into another and per-law scores remain mutually independent. Since the judge runs at a fixed checkpoint with deterministic (greedy) decoding, the same (video, prompt) input yields the same score across runs, eliminating the silent run-to-run drift that closed-source API judges (GPT-4o, GPT-5\,Pro, Gemini-3.1-Pro) exhibit on the same inputs. For each model and each dimension, we report the average score across all applicable generated videos.

\section{Experiments}

\subsection{Experiment Setup}

\paragraph{Training.} In the first stage, we start from Wan2.2-I2V-A14B \cite{wan2025wan}, perform finetuning, and adapt the model to both image-to-video and video-to-video generation with the Diffsynth-Studio framework~ \cite{diffsynth}. Wan2.2-I2V-A14B has two DiTs for different timesteps and we finetune both   DiTs  sequentially. The learning rate is set to  1e-6. With our video-to-video training pipeline, the input video has 17 frames and the ground truth video has 49 frames. 

For the DPO training, with $\text{world\_size}=4$, $\text{micro\_batch}=1$, and gradient accumulation $2$ (effective batch $8$ per optimiser step), $1{,}000$ pairs amount to $125$ optimizer steps per epoch. Training uses AdamW with learning rate $1\times10^{-5}$, micro-batch size $1$ per FSDP rank, and $\beta=100$ in the DPO loss. We train for two epochs ($250$ optimizer steps in total) and select the final-epoch checkpoint without further wrap-around.   We train the model with 16 Nvidia H100 GPUs.

\paragraph{Baselines.} We compare our model with Wan2.2-I2V-A14B \cite{wan2025wan}, Cosmos-14B \cite{agarwal2025cosmos}, LTX-2.3-22B \cite{hacohen2025ltx2}, OmniWeaving \cite{pan2026omniweaving}, Cosmos-2-2B \cite{ali2025world}.  We adopt the video-to-video pipeline to generate videos for comparison. For Wan2.2-I2V-A14B which is an image-to-video model, we adapt the image-to-video pipeline to enable the video-to-video capability.  

\paragraph{Benchmark.} We evaluate the quality of generated videos with 500 random prompts from  VBench~\citep{huang2023vbench}.
Especially, we report the  subject consistency, background consistency, motion smoothness, dynamic degree, aesthetic quality, and imaging quality, from VBench for our generated videos.
All videos are generated with   prompts from VBench, with a resolution of 480p. 

To evaluate the physics enforcement after DPO training,
we evaluate on our TI2V benchmark whose videos are scored on a $1$--$5$ Likert scale by quality-controlled human annotators. We use the benchmark's released automated judge model for evaluation. For PhyWorld and every baseline we compare against, we generate videos on the benchmark's prompt set under each model's default decoding settings, and report the per-dimension and per-law averages produced by the released judge. The same prompt set is used for every model, yielding paired comparisons at the prompt level. We do not modify the benchmark, the prompts, the rubric, or the judge. All of those are taken as released.

\subsection{Main Results}

\begin{table}[t]
\caption{Evaluation results on VBench.  Our PhyWorld achieves an average score of 0.769, demonstrating non-marginal improvements over SOTA baselines with averages scores below 0.756.}
\label{tab:results}
\scalebox{0.88}{
\begin{tabular}{c|cccccc|c}
\toprule
             & \begin{tabular}[c]{@{}c@{}}Subject\\ consistency\end{tabular} & \begin{tabular}[c]{@{}c@{}}Background\\ consistency\end{tabular} & \begin{tabular}[c]{@{}c@{}}Motion\\ smoothness\end{tabular} & \begin{tabular}[c]{@{}c@{}}Dynamic\\ degree\end{tabular} & \begin{tabular}[c]{@{}c@{}}Aesthetic\\ quality\end{tabular} & \begin{tabular}[c]{@{}c@{}}Imaging\\ quality\end{tabular} & Avg. \\ \hline
Cosmos-2-2B & 0.887 & 0.905  & 0.964 &  0.543  & 0.524 &  0.608  & 0.739
             \\ 
OmniWeaving & 0.903  & 0.907   & 0.972  &    0.556 &  0.541 & 0.621  & 0.75
             \\ 
LTX-2.3-22B & 0.894 & 0.918  & 0.982  &  0.549  & 0.532 & 0.626 & 0.751
             \\ 
Cosmos-14B & 0.899 &  0.923 &  0.973 & 0.559  & 0.536 & 0.629  & 0.753
             \\ 
Wan2.2-I2V-A14B & 0.912                                                        & 0.928                                                          &   0.977                                                          &  0.554                                                        & 0.543                                                      & 0.622  &0.756                                                    \\ 
\textbf{PhyWorld (ours)}         & \textbf{0.932}                                                        & \textbf{0.944}                                                           &  \textbf{0.986}                                                           &   \textbf{0.564}                                                       & \textbf{0.555}                                              & \textbf{0.632}          & \textbf{0.769}                                           \\ \bottomrule
\end{tabular}}
\end{table}

\paragraph{Evaluation on Generation Quality.}   We  evaluate the generation quality of our model on the VBench benchmark \cite{huang2023vbench}. As shown in Table~\ref{tab:results}, our method achieves superior performance consistently across different evaluation metrics including subject
consistency, motion smoothness, imaging quality and so on. The average score is 0.769, demonstrating non-marginal improvements over SOTA baselines with averages scores below 0.756.

\paragraph{Evaluation on Physical Faithfulness.}
To evaluate the physical faithfulness, we adopt our judge model to provide scores for the generated videos. We compare our model with SOTA I2V/T2V baselines evaluated on the same $250$ prompts with the same judge model. In Table~\ref{tab:humaneval-250-leaderboard}, we report the   mean score for multiple evaluation metrics including SA/PTV/persistence for general quality, Solid-Body/Fluid/Optical for physical adherence, and Overall as the  mean of the metrics.

We can observe that our PhyWorld reaches $3.09$ Overall, outperforming its own frozen base Wan2.2-I2V-A14B ($2.99$) and  other five  open SOTA models (Cosmos-14B, OmniWeaving, LTX-2.3-22B, Wan2.2-TI2V-5B, LTX-2-19B). 
Our non-marginal improvements over the initial base model   concentrates on the axes that the DPO score $s(v)$ and high-noise training window are designed to move, such as PTV ($+0.10$), Persistence ($+0.15$), and the optical-physics domain ($+0.21$). To complement these aggregate scores with per-prompt evidence, Appendix~\ref{app:comparison} walks through four representative cases (rigid-body, gravity, fluid continuity, and material breakage) where the Wan2.2-I2V-A14B base produces a physically implausible video and our RL-DPO LoRA visibly fixes the violation.

\begin{table}[t]
\centering
\small
\setlength{\tabcolsep}{4pt}
\caption{Scores from our judge model on the 250-prompt humaneval set (1--5 scale, higher is better)  for various models. Semantic alignment (SA), physical-temporal validity (PTV), and persistence are per-video general dimensions. Solid-Body, Fluid, and Optical are pooled over (video, law) units in each domain. Overall = 0.5 $\times$ mean(SA, PTV, Persist.) + 0.5 $\times$ pooled mean over (video, law) units across the three  domains. \textbf{Bold} marks the best score.}
\label{tab:humaneval-250-leaderboard}
\scalebox{1.2}{
\begin{tabular}{l ccc ccc c}
\toprule
& \multicolumn{3}{c}{General Quality $\uparrow$} & \multicolumn{3}{c}{Physics Adherence $\uparrow$} & \\
\cmidrule(lr){2-4} \cmidrule(lr){5-7}
Model & SA & PTV & Persist. & Solid-Body & Fluid & Optical & Overall $\uparrow$ \\
\midrule
    \textbf{PhyWorld (ours)} & \textbf{2.78}   & \textbf{3.07} & \textbf{3.23} & \textbf{2.84}  & \textbf{3.04} & \textbf{3.57}  & \textbf{3.09} \\
    Wan2.2-I2V-A14B & 2.72 & 2.97 & 3.08 & 2.79 & 3.03 & 3.36 & {2.99} \\
    Cosmos-14B & 2.60 & 2.73 & 3.07 & 2.72 & 2.92 & 3.53  & 2.80 \\
    OmniWeaving & 2.68 & 2.73 & 2.92 & 2.71 & 2.99 & 3.13 & 2.78 \\
    LTX-2.3-22B & 2.63 & 2.79 & 2.91 & 2.55 & 3.02 & 3.21 & 2.72 \\
    Wan2.2-TI2V-5B & 2.48 & 2.70 & 2.76 & 2.61 & 3.01 & 3.45 & 2.68 \\
    LTX-2-19B & 2.50 & 2.62 & 2.79 & 2.49 & 3.01 & 3.09 & 2.62 \\
\bottomrule
\end{tabular}}
\end{table}

\subsection{Visual Demonstration}

\begin{figure}[t]
    \centering
    \includegraphics[width=1.0\textwidth]{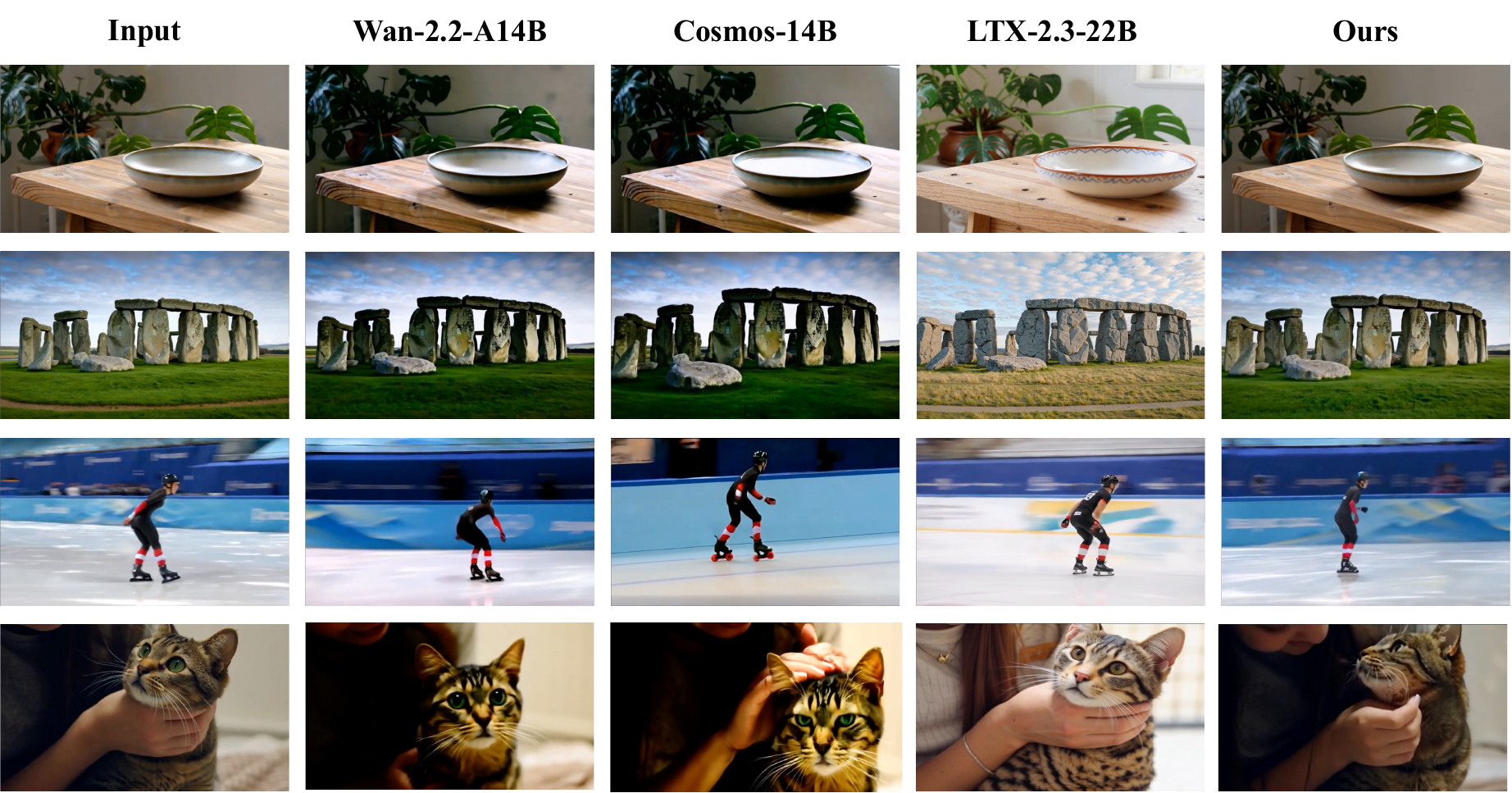}
    \caption{Visual comparison with baselines.  PhyWorld  generates videos with superior physical consistency. The SOTA baseline models suffer from artifacts or flaws such as  color shift or background change, which are not physically consistent.}
    \label{fig:visual}
\end{figure}

We demonstrate the visualization of PhyWorld and compare with baselines including Wan-2.2-I2V-A14B, Cosmos-14B, and LTX-2.3-22B.  We adopt the video-to-video pipeline for all models. For   Wan-2.2-I2V-A14B, we update the image-to-video pipeline to enable the video-to-video capability. As shown in Figure~\ref{fig:visual} with images captured  from the generated videos, PhyWorld can generate  videos with superior physical consistency.  The   baselines suffer from artifacts or flaws such as  color shift or background change. More visual examples are shown in Appendix~\ref{app:sec:visual_example}.

\section{Conclusion}
In this work, we present PhyWorld, a two-stage world model framework for generating physically consistent and faithful video continuations. By combining flow matching fine-tuning for temporal coherence with Direct Preference Optimization for physics enforcement, our approach addresses the two fundamental limitations of existing video generation models: physical inconsistency across frames and the absence of principled physics grounding. We evaluate PhyWorld on our dedicated benchmark for evaluating physics faithfulness in generated video, against which PhyWorld demonstrates superior performance. 
We hope this work advances the development of scalable, realistic world simulators for Physical AI training and inspires further research into physics-aware video generation.  
PhyWorld is post-trained from Wan2.2-I2V-A14B and inherits its data biases and failure modes. The transfer of policies trained inside PhyWorld is left to future work.

{
\small
\bibliographystyle{unsrt}
\bibliography{ref}
}

\newpage
\appendix

\section{Benchmarks for Physical Faithfulness} \label{app:sec:failure}

Current physics-evaluation pipelines for video generation suffer from a compounding set of design flaws that together leave their scores only weakly coupled to actual physical correctness. The most fundamental issue is prompt-level: benchmarks such as VideoPhy-2, Physics-IQ, and WorldModelBench rely on implicit prompts that describe a scene without specifying its expected  physical outcome, so an evaluator can only judge whether the video looks plausible rather than whether the physics is correct — "a  wooden block slides across a table" need never reach the edge or fall, and a physically wrong but visually coherent generation can still earn a high score with the violation undetected. 

The second flaw is scoring-level: nearly  every prior benchmark reports a single holistic physics aggregate, or at best binary per-event judgments on a <5 level scale, which collapses heterogeneous failure modes- a model that obeys optical reflection while systematically violating momentum  conservation receives the same score as one whose error pattern is the inverse - and prevents per-law diagnostics.

Pixel-reference metrics such as Physics-IQ's S-IoU, ST-IoU, and MSE inherit a  different but equally fatal pathology: they assume a unique ground-truth trajectory and penalize cinematographic camera motion, even though many physical outcomes — collision rebound angles, fluid splashes — are inherently stochastic and admit multiple  equally valid trajectories that can yield drastically different IoU values; the reference-based design moreover cannot be extended  to the text-to-video setting at all, since no ground-truth video exists by construction. 

\begin{figure}[h]
    \centering
    \includegraphics[width=1.0\textwidth]{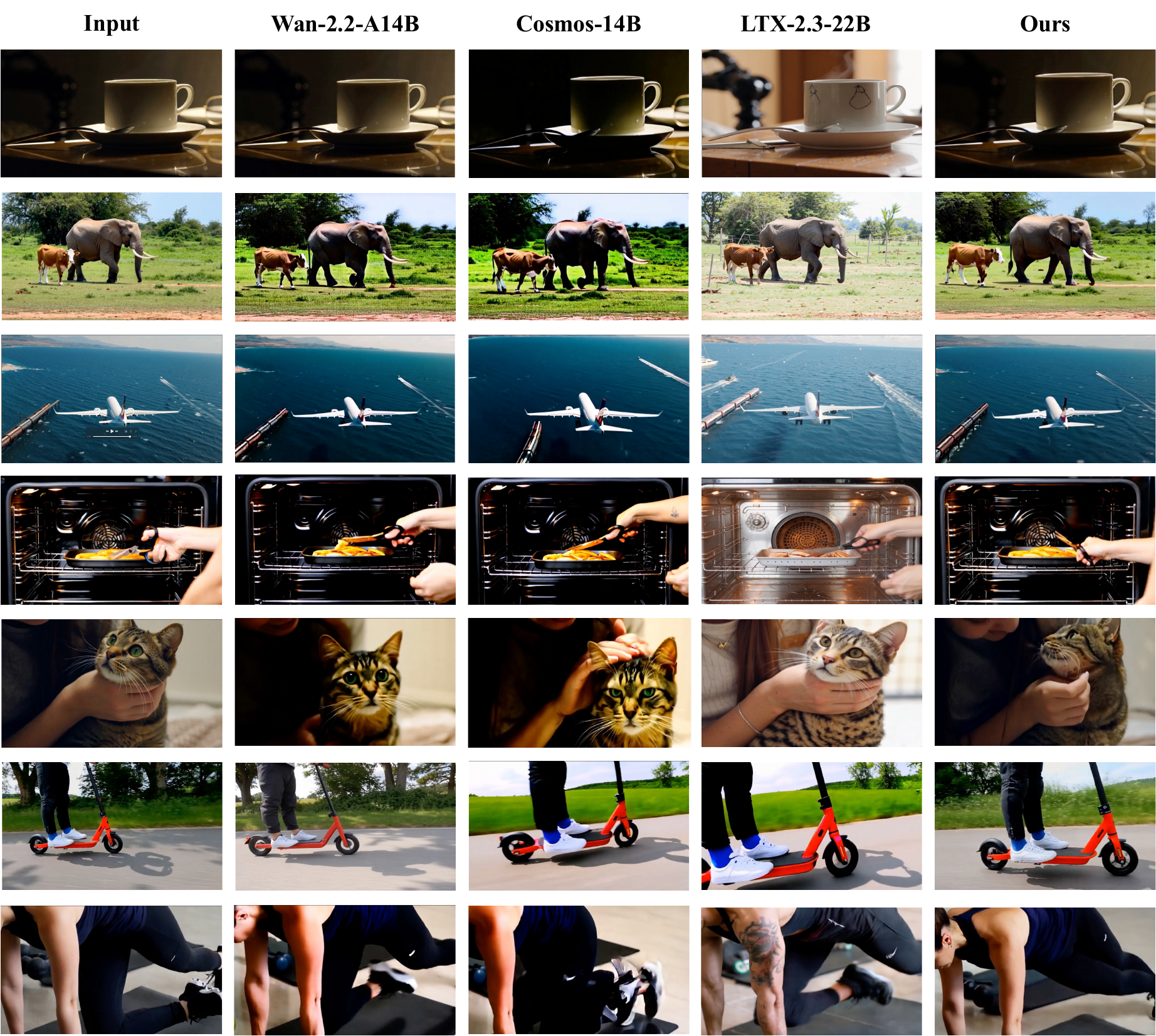}
    \caption{Visual comparison with baselines.  PhyWorld  generates videos with superior physical consistency, while   baselines suffer from artifacts or flaws such as  color shift or background change.}
    \label{app:fig:visual1}
\end{figure}

Many benchmarks rely on closed-source evaluators (including GPT-4o, GPT-5, and Gemini-3.1-Pro) to provide physical evaluations. However, it introduces compounding reliability concerns. The behavior of these models is subject to silent drift across     model versions, weights, and API   changes outside benchmark users' control.  The empirical inspection reveals systematic failure modes in specific physical domains. In particular, Gemini-3.1-Pro has been observed to classify visibly dynamic fluid motion as entirely static, and to produce markedly unstable scores on shadow and reflection phenomena.

\section{More Details for our Benchmark for Physics Alignment} \label{app:sec:more_detail}

\begin{figure}[t]
    \centering
    \includegraphics[width=1.0\textwidth]{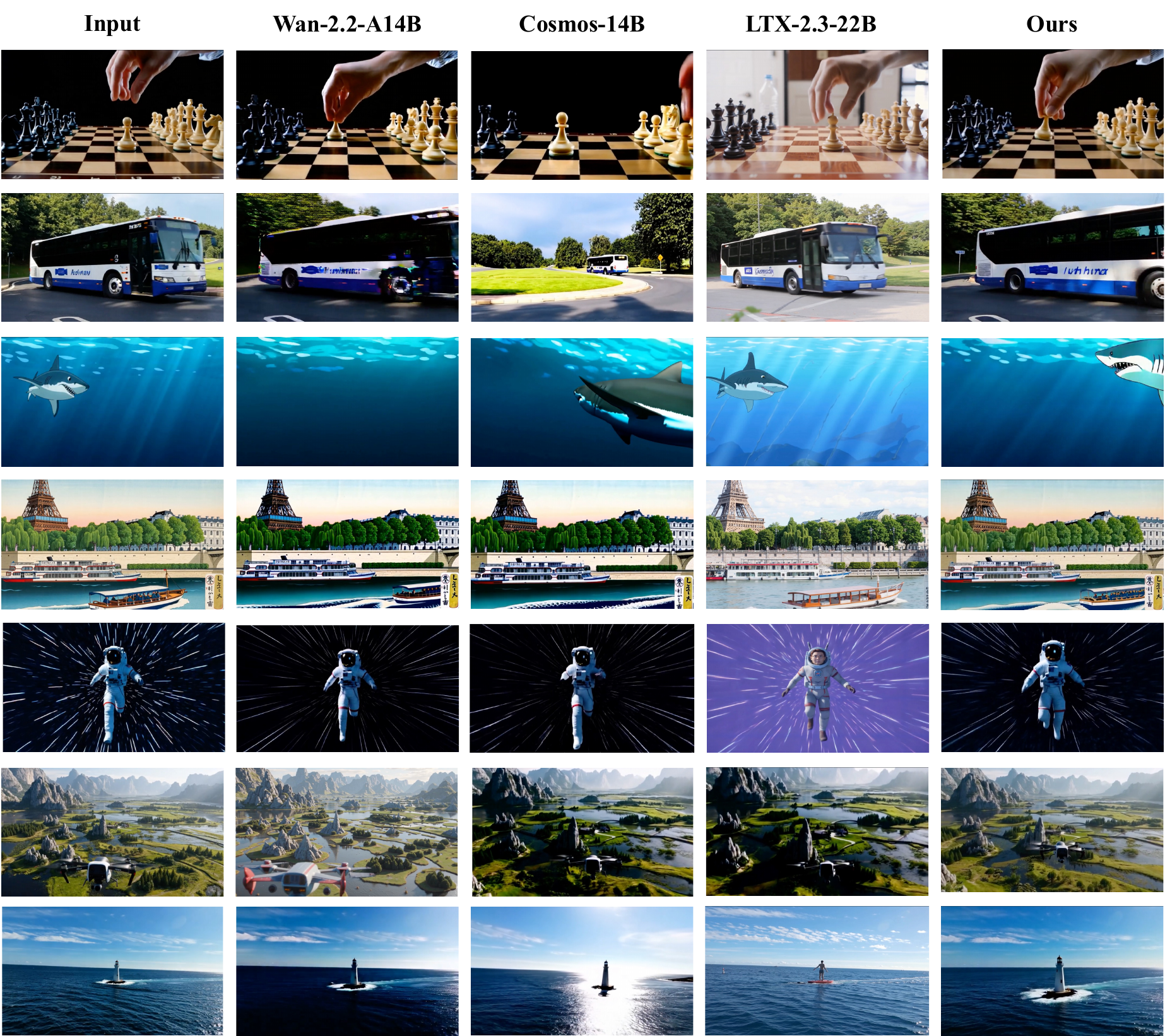}
    \caption{Visual comparison with baselines.  PhyWorld  generates videos with superior physical consistency, while   baselines suffer from artifacts or flaws such as  color shift or background change.}
    \label{app:fig:visual2}
\end{figure}

\paragraph{Annotators and onboarding.} Around 350 annotators are recruited through a web-based platform and pass through a fixed three-stage onboarding before rating anything: a consent / study-overview page, a structured demographic questionnaire (age, gender, education, major), and a  training module of three calibration videos spanning different levels of physical realism so non-expert raters acquire a shared  scoring reference. 

\paragraph{Per-rater protocol.} Each rater receives a randomly assigned subset of the 2,000 videos (250 prompts × 8  generators) with the within-subset order independently randomized to neutralize ordering effects, must play every video at least  once before submitting, and scores it on three general 1–5 Likert dimensions — semantic alignment (SA), physical-temporal validity   (PTV), persistence — plus up to three of the thirteen applicable physics laws; per-video aggregates use the median across raters   rather than the mean to limit outlier sensitivity. Quality control. Raw annotations (5,796 submissions, 37.4K labels) pass through  a two-round filter that combines four mutually independent signals — score constancy (flag std < 0.3, including the std = 0  raters who give the same number to everything), per-video copy-paste rate (identical scores across dimensions inside one video even when cross-video std > 0, which constancy alone cannot catch), peer mean absolute error on shared videos, and behavioral telemetry (page stay time, video play count) — and an annotator is removed only when several signals are simultaneously abnormal,  retaining 352 of 459 raters and yielding a cleaned corpus of 4,576 annotations and 29.5K labels.

\begin{figure}[t]
\centering
\includegraphics[width=\linewidth,height=\textheight,keepaspectratio]{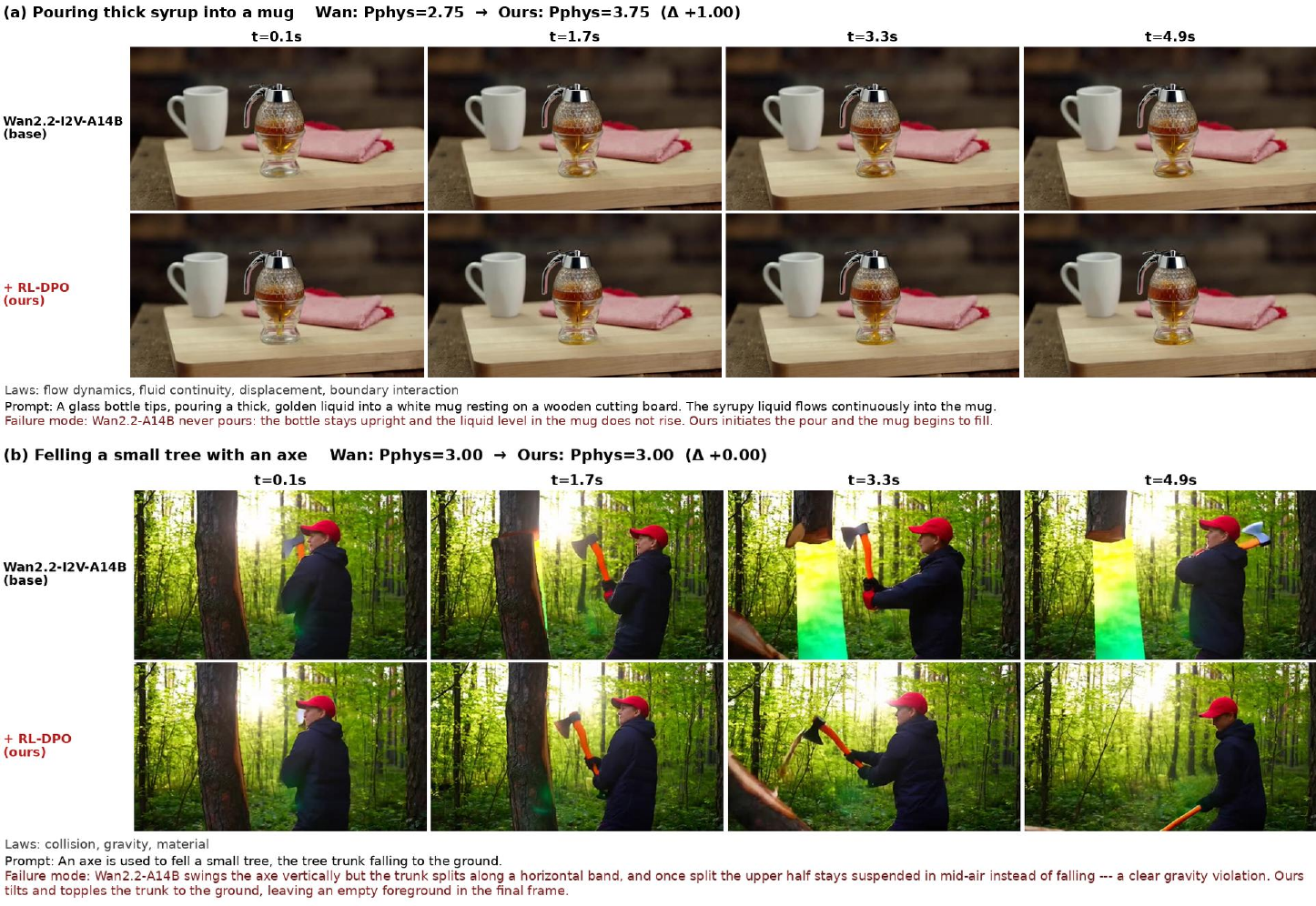}
\caption{Per-prompt comparison (1/2): rigid-body and gravity violations. (a) Pouring syrup --- Wan2.2-A14B never initiates the pour and the mug stays empty; ours starts the pour and the mug begins to fill. (b) Felling a tree --- Wan2.2-A14B swings the axe \emph{vertically} but the trunk splits along a \emph{horizontal} band, after which the upper half stays suspended in mid-air rather than falling under gravity; ours tilts and topples the trunk to the ground.}
\label{fig:comparison_ab}
\end{figure}

\paragraph{From annotations to training corpus.} The judge model is trained on a subset of this same pool, namely 8 generators $\times$ 93 prompts   about 13K (video, dimension) records, partitioned at the (generator, prompt) pair level with random seed 42 into four disjoint splits   — TRAIN (seen × seen), TEST-PROMPT (seen generators × 12 unseen prompts), TEST-MODEL (the fully held-out veo-3.1 generator × seen   prompts), and TEST-BOTH (unseen × unseen) — so generalization is measurable along both axes; because the partition is at the  (generator, prompt) level rather than the law level, all 13 physics laws appear in roughly equal proportions across splits, and a  further 5\% of videos inside the training pool (video-disjoint but otherwise drawn from the same prompts and generators) is held in   as a validation set used solely for early stopping by eval loss. Training samples. Each sample is a triple of (a) video frames  sampled at 4 fps with the short side resized to 360 and capped at FPS MAX FRAMES = 12, IMAGE MAX TOKEN NUM = 1024 per frame, and  VIDEO MAX TOKEN NUM = 1024 over the whole clip, (b) the augmented prompt containing the expected physical outcome — bit-identical  to the prompt the judge sees at inference time, so train/inference distribution drift is zero — and (c) a structured JSON score  block consisting of a single key (the target dimension or physical law) and a single integer 1–5 value derived from the human  annotations.

\begin{figure}[t]
\centering
\includegraphics[width=\linewidth,height=\textheight,keepaspectratio]{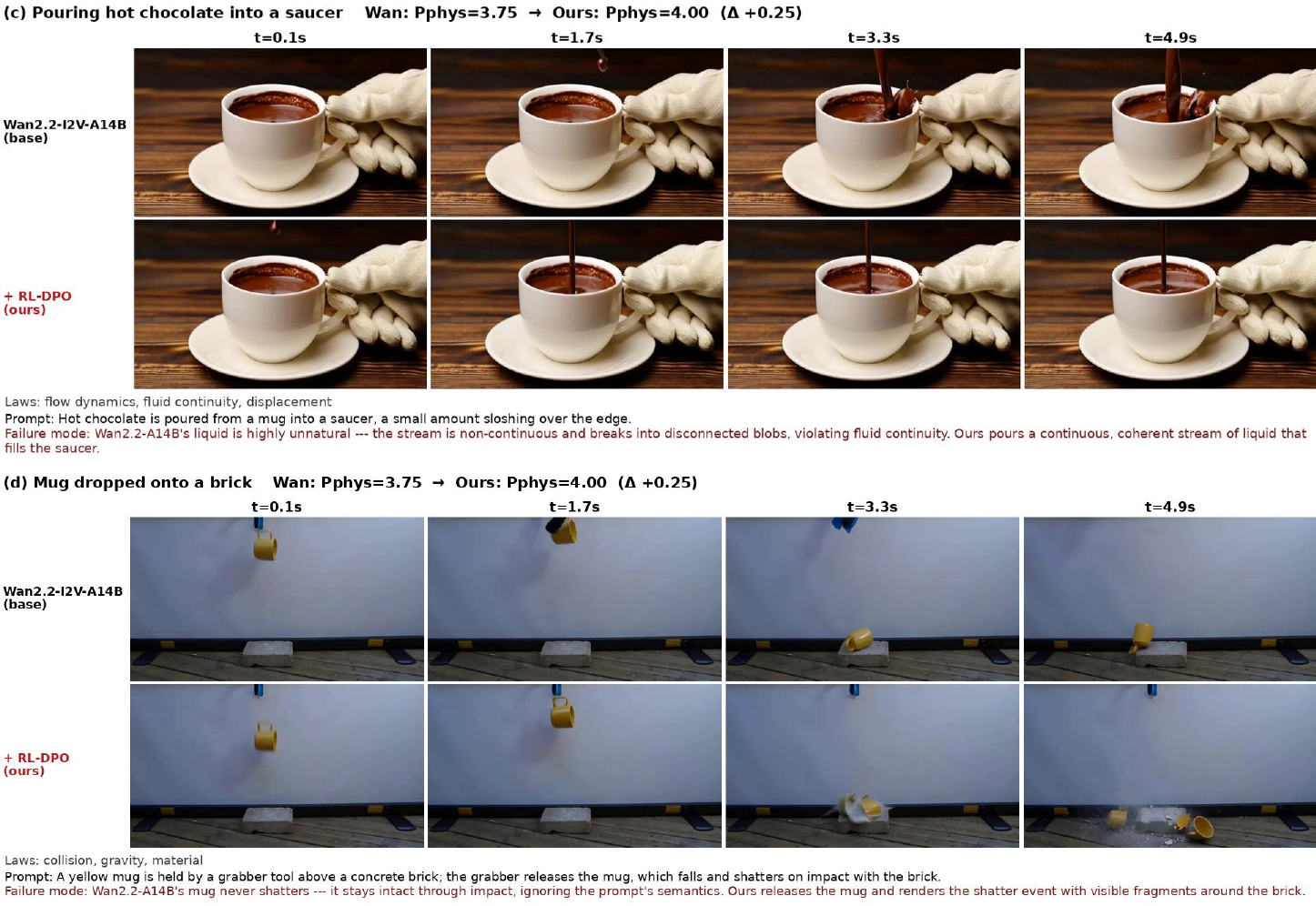}
\caption{Per-prompt comparison (2/2): fluid continuity and material/breakage. (c) Pouring hot chocolate --- Wan2.2-A14B's stream is non-continuous and breaks into disconnected blobs, violating fluid continuity; ours produces a continuous coherent stream that fills the saucer. (d) Dropping a mug onto a brick --- Wan2.2-A14B's mug never shatters and ignores the prompt's semantics; ours releases the mug and renders the shatter event with visible fragments around the brick.}
\label{fig:comparison_cd}
\end{figure}

The base model is Qwen3.5-9B — chosen because it fits a single GPU (TP = 1), supports video input  natively, and after SFT closes the gap to the 27B variant (which differs only by switching DeepSpeed from ZeRO-2 to ZeRO-3) — and LoRA is applied at rank 32, alpha 64, dropout 0.05 across all linear layers (q/k/v/o plus MLP gate/up/down) of the LLM; the visual  encoder (ViT) is frozen to preserve visual understanding, the aligner is left trainable but on a smaller learning rate (2e-6  versus 1e-4 for the LoRA adapters), and the optimizer runs in bfloat16 with DeepSpeed ZeRO-2 (ZeRO-3 for the 27B ablation).  Training is one epoch with per-device batch size 1 and gradient accumulation 8 (effective batch 8), max sequence length 8192,  warmup ratio 0.05, gradient clipping at 1.0, evaluation every 100 steps, and the checkpoint with the lowest eval loss is loaded as   final. Inference and per-metric supervision. At both training and inference the judge is queried in per-metric mode: each  evaluation dimension — SA, PTV, persistence, and every applicable physics law — is scored in a separate inference call, and the  assistant target on each call is restricted to a single JSON key-value pair, matching the inference-time output schema exactly so  the model never learns to imitate free-form rationales and reasoning across dimensions cannot contaminate one another. The  combined effect is that judge accuracy on the four held-out splits is bounded by the quality of the human label distribution  rather than by any architectural choice — exactly the property required for a benchmark whose primary contribution is  reproducibility.

\section{Visual Demonstrations} \label{app:sec:visual_example}

We demonstrate more visual comparison in Figure~\ref{app:fig:visual1} and \ref{app:fig:visual2}.

\section{Per-prompt comparison: Wan2.2-I2V-A14B base vs.\ our RL-DPO LoRA}
\label{app:comparison}

To complement the leaderboard in Table~\ref{tab:humaneval-250-leaderboard}, we walk through four prompts on which the open-source base model \emph{Wan2.2-I2V-A14B} \citep{wan2025wan} produces a physically implausible video, and on which our RL-DPO LoRA on the same backbone visibly fixes the violation. In Figure~\ref{fig:comparison_ab} and \ref{fig:comparison_cd}, each row is a stitched composite: the upper strip is the Wan2.2-I2V-A14B base output and the lower strip is our checkpoint, both decoded from the same first frame and prompt at the same four timestamps ($t \approx 0.1, 1.7, 3.3, 4.9$\,s of the 5\,s, 16\,fps clip). The case header reports our judge model's physics score $P_{\rm phys}$ (1--5) for each model and the gain $\Delta$.



\end{document}